\documentclass[nohyperref]{article}

\usepackage{microtype}
\usepackage{graphicx}
\usepackage{subfigure}
\usepackage{booktabs} 

\usepackage{hyperref}

\usepackage[accepted]{icml2022}

\usepackage{amsmath}
\usepackage{amssymb}
\usepackage{mathtools}
\usepackage{amsthm}
\usepackage{amsmath}

\DeclareMathOperator*{\argmin}{arg\,min}
\usepackage[capitalize,noabbrev]{cleveref}

\theoremstyle{plain}
\newtheorem{theorem}{Theorem}[section]

\theoremstyle{definition}
\newtheorem{definition}[theorem]{Definition}

\theoremstyle{remark}

\usepackage[textsize=tiny]{todonotes}

\icmltitlerunning{Disentangled Federated Learning}

\begin{document}

\twocolumn[
\icmltitle{Disentangled Federated Learning for Tackling Attributes Skew \\
via Invariant Aggregation and Diversity Transferring}



\icmlsetsymbol{equal}{*}

\begin{icmlauthorlist}
\icmlauthor{Zhengquan Luo}{ustc,casia}
\icmlauthor{Yunlong Wang}{casia}
\icmlauthor{Zilei Wang}{ustc}
\icmlauthor{Zhenan Sun}{casia}
\icmlauthor{Tieniu Tan}{casia}
\end{icmlauthorlist}

\icmlaffiliation{ustc}{University of Science and Technology of China (USTC)}
\icmlaffiliation{casia}{Institute of Automation, Chinese Academy of Sciences (CASIA)}

\icmlcorrespondingauthor{Zhengquan Luo}{zhengquan.luo@cripac.ia.ac.cn}
\icmlcorrespondingauthor{Yunlong Wang}{yunlong.wang@cripac.ia.ac.cn}
\icmlcorrespondingauthor{Zilei Wang}{zlwang@ustc.edu.cn}
\icmlcorrespondingauthor{Zhenan Sun}{znsun@nlpr.ia.ac.cn}
\icmlcorrespondingauthor{Tieniu Tan}{tnt@nlpr.ia.ac.cn}

\icmlkeywords{Machine Learning, ICML}

\vskip 0.3in
]



\printAffiliationsAndNotice{}  

\begin{abstract}

Attributes skew hinders the current federated learning (FL) frameworks from consistent optimization directions among the clients, which inevitably leads to performance reduction and unstable convergence. The core problems lie in that: 1) Domain-specific attributes, which are non-causal and only locally valid, are indeliberately mixed into global aggregation. 2) The one-stage optimizations of entangled attributes cannot simultaneously satisfy two conflicting objectives, i.e., generalization and personalization. To cope with these, we proposed disentangled federated learning (DFL) to disentangle the domain-specific and cross-invariant attributes into two complementary branches, which are trained by the proposed alternating local-global optimization independently. Importantly, convergence analysis proves that the FL system can be stably converged even if incomplete client models participate in the global aggregation, which greatly expands the application scope of FL. Extensive experiments verify that DFL facilitates FL with higher performance, better interpretability, and faster convergence rate, compared with SOTA FL methods on both manually synthesized and realistic attributes skew datasets.

\end{abstract}

\begin{figure}[ht]
\vskip 0.2in
\begin{center}
\includegraphics[width=0.9\columnwidth]{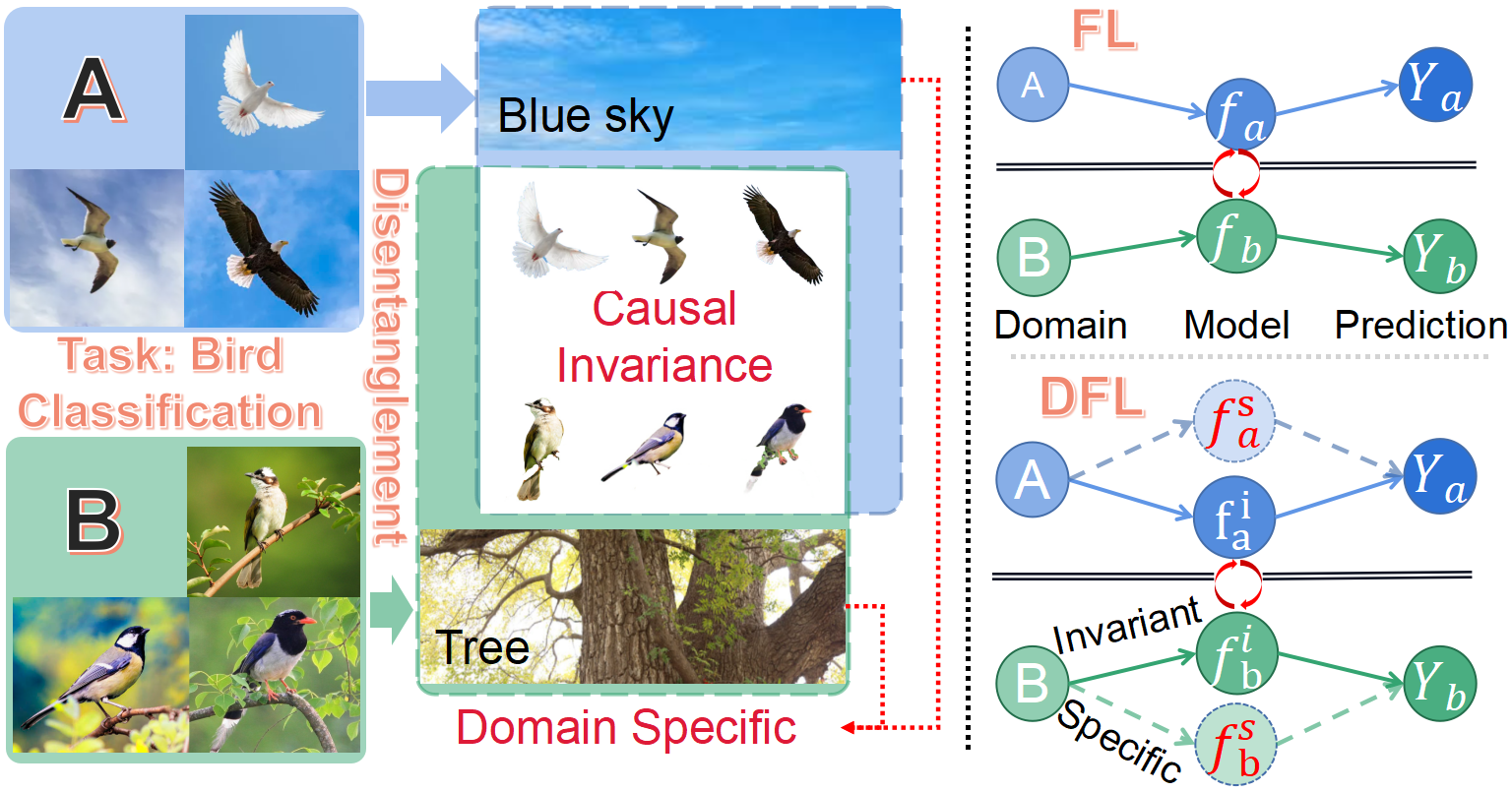}
\vskip -0.05in
\caption{Taking bird classification as an example to illustrate the principle of DFL. The domain-specific attributes, such as the blue sky or trees, are stripped out and contributed locally. The invariant aggregation employs the partial client model, which only focuses on the invariant and causal attributes, like the bird itself. The two-branch local model is employed in DFL to replace the single-branch in FL, which shows in the right part.}
\label{fig:principle}
\end{center}
\vskip -0.2in
\end{figure}

\section{Introduction}
\label{sec:intro}
Federated learning~\cite{mcmahan2017communication}, as a decentralized and privacy-preserving machine learning framework, aims to build a client-server system that can adapt to different distributions without access to local raw data. A key challenge lies in the Non-i.i.d factors between clients' data domains. The huge distribution shift comes from the diverse contextual information across devices/locations, which causes suboptimal or detrimental performance. In addition, the aggregated global model could be unstable and even nonconvergent due to the different optimization directions of client models. To mitigate this problem, a plethora of techniques are introduced to adjust the system, such as local model adaptation, global reweighting aggregation, and optimization direction correction. These methods make FL rapidly adapt to the local distributions and corresponding tasks. Although effective, the basic problem still exists. Especially, the attributes skew, as one challenging factor of Non-i.i.d, indicates the scenarios in which the representation distribution across attributes on each client is different from each other. The domain-specific attributes are inevitably extracted by the single-branch client model and mixed into the global aggregation. These attributes are decision-correlated but not causal, which is only locally valid. It leads to essential differences in the optimization directions of client models. Taking bird classification as an example in the left part of \cref{fig:principle}. The attributes of the bird itself and the blue sky are extracted by the client network simultaneously by the local extractor, which both contribute to the task decision in the flying birds' domain. Unfortunately, this blue sky as domain-specific attributes will be infused into the global server by model aggregation. It causes performance degradation to the domain where birds all sit on tree branches. Because there is no blue sky in this data domain. This phenomenon of attributes skew distribution is widespread, which causes concerns about the robustness and the trustworthiness of the FL system. 

To mitigate this negative transfer\footnote{Negative transfer, i.e., leveraging other clients' knowledge undesirably reduces the learning performance of the local client~\cite{wang2019characterizing}.} caused by attributes skew, we proposed disentangled federated learning (DFL). The motivation is to spin off the domain-specific attributes from model aggregation. However, the single-branch-based client model of the traditional FL framework cannot support DFL. The reason is that the specific and invariant attributes are entangled and extracted by a single extractor. Although these two attributes both contribute to the task decision locally, the huge differences are that: 1) invariant attributes are intrinsic and causal, which is cross-domain generic; 2) specific attributes are only locally valid, which may bring performance degradation to other domains. Thus, DFL applied two complementary branches client model, which are presented in the right part of \cref{fig:principle}. These two attributes are disentangled by mutual information (MI) constraints and focused by two branches respectively. 1) Domain-specific branch is only trained locally. 2) Domain-invariant branch is applied for global aggregation. Except for redesigning the local model, the other two innovations are proposed as invariant aggregation and diversity transferring.

{\bf Invariant aggregation} is proposed to employ the local invariant branch for global model aggregation. The MI maximization between the local invariant branch and the global invariant model restricts clients' optimizations in the same direction. The combination of the invariant aggregation and MI constraint drives the local invariant branch to focus on the intrinsic and causal attributes (e.g. bird itself). In the theory of ~\cite{scholkopf2012causal,peters2016causal}, these cross-domain invariant attributes can provide transferable and reliable knowledge, which leads to proper domain adaptation and lighter catastrophic forgetting. In addition, MI minimization is employed to disentangle the invariant and specific locally. Although these domain-specific attributes are dropped out from model aggregation, they can still contribute to the final decision (e.g., things that fly in the blue sky are most likely birds). It is not wise to throw these task-correlated domain-specific attributes away directly. {\bf Diversity transferring} is proposed to make full use of these attributes, enhancing the diversity of representation. In theoretical analysis, this diversity augmentation has proved to mitigate overfitting and correct inaccurate distributions~\cite{shorten2019survey}. Specifically, the diverse representations are extracted by combining the local invariant extractor and transferred specific extractors. Thus, the local client model is forced to pay attention to these tailed attributes, which the local extractor may ignore but has a decision contribution (e.g. trees may appear in the blue-sky domain).

We emphasize that two-stage training in an alternating manner is an essential innovation because it changes the optimization purpose of FL. Compared with one-stage global optimization, alternating optimization easily finds the optimal solution of the global invariant model based on multiple local optimal points. Specifically, part of the optimization process is separated into local clients and is personalized trained. It provides both generalized adaption and personalized performance. Besides, the theoretical analysis proves that DFL is convergent even if incomplete client models participate in the global aggregation, based on the bounded gradient of the remaining model part. As far as we know, this work is the first to provide the convergence guarantee of partial aggregation. With sufficient theoretical guarantees, we design the following experiments: 1) {\bf Clarification experiments} demonstrate the unstable convergence and performance degradation with the introduction of manually synthesized attributes skew in Colored-MNIST~\cite{arjovsky2019invariant}; 2) {\bf Verification experiments} verify the superiority of DFL on convergence rate and classification accuracy, compared with other SOTA personalized FL methods. In addition, the ablation study proves the complementary effectiveness of invariant aggregation and diversity transferring; 3) {\bf Application experiments} on DomainNet~\cite{peng2019moment} and Office-Caltech~\cite{gong2012geodesic} point out that DFL can adapt to the realistic attributes skew. The accuracies are all greatly improved in different backbones. The visualization on DomainNet verifies that the specific and invariant attributes can be successfully disentangled,  proving the interpretability improvement by DFL. In summary, the main contributions of the paper are as follows:

\begin{itemize}
    \item Disentangled federated learning (DFL) is proposed to overcome the attributes skew essentially, which spins off the domain-specific attributes from model aggregation. 
    \item Theory deduction proves the convergence analysis of invariant aggregation based on the bounded local-specific gradient.
   \item Invariant aggregation and diversity transferring are proposed to correct the optimization directions and augment representation diversity.
   \item Alternating local-global optimization is proposed to simultaneously meet generalized adaption and personalized performance.
\end{itemize}

\section{Related Work}
\label{sec:Related}
{\bf Federated learning} research with clients' Non-i.i.d distributions aims to enhance the stability and convergence of FL system which suffers from distribution shift caused by Non-i.i.d factors. Several kinds methods have been proposed from different perspectives: 1) \textbf{Local model adaptation} is proposed to adjust the local model for heterogeneous distributions, such as fine-tuning~\cite{wang2019federated}, Meta-learning-based different initialization~\cite{fallah2020personalized,chen2018federated}, and personalized prediction layers~\cite{arivazhagan2019federated,liang2020think}, etc. 2) \textbf{Global reweighting aggregation} is introduced to employ different global aggregated weights for each client. The weights can be based on data distribution similarity~\cite{huang2021personalized}, local model contribution differences~\cite{zhang2020personalized}, and same consensus focus~\cite{feng2020kd3a}, etc. 3)  \textbf{Optimization direction correction} is proposed to mitigate the gap between local and global models. The additional constraints are introduced to the loss function or optimization, such as regularization terms~\cite{hanzely2020federated}, proximal terms~\cite{li2018federated}, gradient correction~\cite{acar2021debiasing}, Moreau Envelopes~\cite{dinh2020personalized}, attentive message~\cite{huang2021personalized}, control variate~\cite{karimireddy2020scaffold}, etc. Some adversarial-based methods are proposed, such as domain adaptation~\cite{peng 2019 federated} and debiasing ~\cite{hong2021federated}.

{\bf MI-based disentanglement} aims at interpreting underlying interaction factors. MI can measure the degree of interdependence between any two variables. MI can be applied to 1) quantify the separation of distributions in learning binary hash codes by ~\cite{cakir2017mihash,cakir2019hashing}; 2) perform unsupervised learning~\cite{hjelm2018learning}; 3) disentangle the specific features in advertising training by ~\cite{peng2019domain}. In addition, the MI maximization techniques are expended by ~\cite{belghazi2018mutual} via a neural network to estimate MI between two random variables. In this work, MI is employed for two intentions: 1) disentanglement of local specific and invariant attributes; 2) similarity enhancement of local and global invariant model.

\section{Disentangled Federated Learning}
\label{sec:framework}
Rethinking the limitation of single-branch model sharing, we proposed alternating training, which changes the optimization purpose in \cref{sec:Definition}. Two innovations, invariant aggregation and diversity transferring, are proposed to mitigate the attributes skew, and the MI-based disentanglement technique is introduced in \cref{sec:framework}. DFL supported by these methods disentangled the specific and invariant attributes. First, we provide the convergence analysis of DFL in \cref{sec:Convergence Analysis}, which proves that the FL system is convergent even if the shared model is incomplete.

\subsection{Definition of Alternating Optimization}
\label{sec:Definition}
\begin{figure}[t]
\vskip 0.2in
\begin{center}
\includegraphics[width=0.9\linewidth]{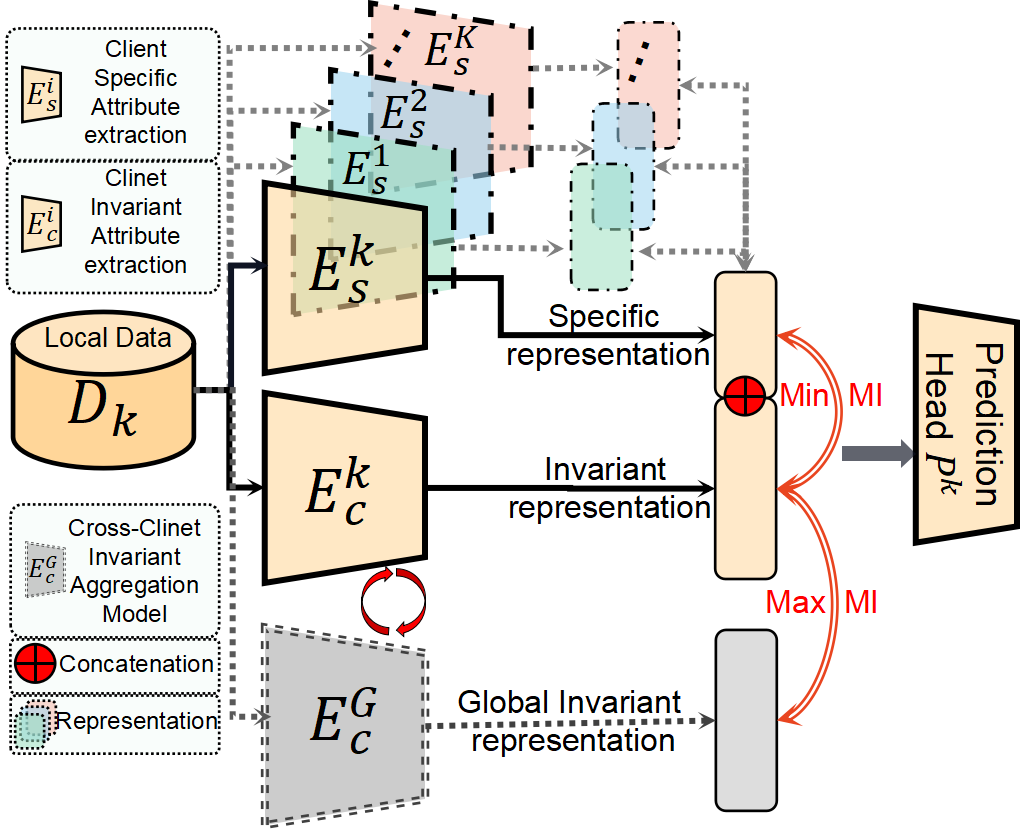}
\vskip -0.1in
\caption{The framework of DFL.}
\label{fig:framework}
\end{center}
\vskip -0.2in
\end{figure}

The task of FL can be defined as follows:
\begin{equation}
\small
\begin{aligned}
\underset{\omega}{min}\left \{ f\left ( \omega  \right ):=\frac{1}{N}\sum_{k=1}^{N}h_{k}(\omega) \right \}
\end{aligned}
\label{eq:FLtask}
\end{equation}
The optimization of a local client is to minimize the loss of each client $k\in \left | K \right |$. 
The key problem of this purpose is the Non-i.i.d dilemma in FL. Specifically, the personalized loss minimization pushes the client parameter converging to the local optimal point following the client data distribution $D_{k}^{*}$. However, the model aggregation of FL drives server model gradient descent toward the global direction, which has a huge direction shift from the local client's optimization $D_{i}^{*}\neq D_{j}^{*} \neq D^{*}, 1\leqslant  i\neq j\leqslant K$. 
The reason is that the domain-specific and invariant attributes are entangled and indiscriminately extracted by the single-branch local extractor. It leads to generalization problems. The MI-based disentanglement is introduced to disentangle these two attributes into two local branches, which are optimized independently. The client network is divided as representation extractors, i.e., the invariant branch $E_{c}^{k}$ and the specific branch $E_{s}^{k}$. The prediction module $P^{k}$ takes the concatenated representations from these two branches as input for final decisions. The entire model framework is shown in \cref{fig:framework}.

It should be emphasized that attribute disentanglement is leveraged to break through the limitation of one-stage global optimization of traditional FL methods. One-stage global optimization strives to find an optimal solution that simultaneously meets two conflicting objectives: generalized adaption and personalized performance. However, such efforts are usually in vain. To overcome this problem, we proposed two-stage alternating optimization. Specifically, only the invariant extraction branch of each client participates in the global model aggregation. The specific branches are optimized locally. The optimization purpose of DFL is changed to:

\begin{equation}
\begin{aligned}
\underset{\omega_{c}}{min}\begin{Bmatrix}
f(\omega_{c}):=\frac{1}{N}\sum_{k=1}^{N}\underset{\omega _{k,s}}{min}h_{k}(\omega _{i})\\ 
\omega _{i}=M(\omega _{c},\omega _{k,s})=P_{c}\omega _{c}+P_{s}\omega _{k,s}
\end{Bmatrix}
\end{aligned}
\label{eq:new_MIFL_question_define}
\end{equation}
where $M$ represents the model combination of two branches, $\omega _{c}$ is the parameter of the aggregated invariant model. $h_{k}$ is local loss function. $P_{c}$ and $P_{s}$ are the weighting vectors of an invariant and specific model in local combinations. The minimized local specific branch model is defined as $\omega _{k,s}^{*}$, which satisfies the condition as:
\begin{equation}
\begin{aligned}
\omega _{k,s}^{*}=\underset{\omega _{k,s}}{\argmin}h_{k}(M(\omega _{c},\omega _{k,s}))
\end{aligned}
\label{eq:minimized_local_specific}
\end{equation}
Compared with the one-stage methods, the optimization of local specific branches in DFL is locally trained at first. Then, the entire local model parameter is $\omega _{k}^{*}=M(\omega _{c},\omega _{k,s}^{*})$ based on multiple optimal points $\omega _{k,s}^{*}$ of local specific branches. The global purpose turned to:
\begin{equation}
\begin{aligned}
\underset{\omega_{c}}{min}\left \{ f(\omega_{c}):=\frac{1}{N}\sum_{k=1}^{N}h_{k}(\omega _{k}^{*}) \right \}
\end{aligned}
\label{eq:new_MIFL_question_define_easy}
\end{equation}
which aims to find the global optimal solution of the invariant branch $\omega _{c}$. 
In summary, one-stage optimization is divided into two alternating local and global parts. The advantages are: 1) Finding the optimal global invariant aggregated model with better generalization is easier. 2) The specific branch is optimized locally, which provides personalized adaption. 3) The stability of convergence is enhanced by alternating local-global optimization.

\subsection{Framework}
\label{sec:Framework}
{\bf Representation disentanglement:} For effective attributes disentanglement, the MI-based disentanglement technique~\cite{belghazi2018mutual} is introduced to disentangle the local extractors. MI has been applied in two aspects: 1 ) MI maximization between local invariant and global invariant branches enhance the cross-domain similarity. 2) MI minimization between local invariant and local specific branches disentangles the mixed attributes.
Thus, the adversarial objective function of the client $k$ model is defined as:
\begin{equation}
\begin{aligned}
L_{MI}^{k}:= I_{s}(E_{s}^{k}(x^{k}),E_{c}^{k}(x^{k}))-I_{c}(E_{c}^{k}(x^{k}),E_{c}^{G}(x^{k}))
\end{aligned}
\label{eq:client_MI}
\end{equation}
Deep InfoMax proposed by~\cite{hjelm2018learning} is employed as MI estimation, which is based on Jensen-Shannon estimator, i.e.$I^{JSD}(X,Z)=D_{JS}(\mathbb{P}_{XZ}||\mathbb{P}_{X}\mathbb{P}_{Z})$.

{\bf Invariant aggregation} is proposed to apply the weighted averaging to the invariant extractors from selected clients as:
\begin{equation}
\begin{aligned}
\mathbb{E}_{c}^{G}=\omega_{k}\mathbb{E}_{c}^{k}=\sum _{k=1}^{K}\frac{n_{k}}{N}\mathbb{E}_{c}^{k}
\end{aligned}
\label{eq:invariant_aggregation}
\end{equation}
At the beginning of optimization, the aggregated model may contain inadvertently mixed local domain-specific attributes. When enough clients participate, one domain-specific attribute is reduced to $\frac{n_{k}}{N}$ in the aggregated model by model averaging. Thus, for better specific/invariant disentanglement, parameters of the aggregated model are frozen. MI maximization drives the local invariant branch close to global aggregation, which abandons the domain-specific attributes and pays more attention to the cross-domain invariant attributes. After that, the optimal local invariant branch contributes to the global invariant aggregation.

{\bf Diversity transferring} With the successful disentanglement, the domain-specific attributes are stripped out from the model aggregation. Most pioneering FL methods directly ignored these attributes, which causes substantial waste. The domain-specific attributes contain not only the personalized requirements of local clients but also the diversities of data distributions. Thus, diversity transferring is proposed to augment the local representation sets as:
\begin{equation}
\begin{aligned}
\left \{ R_{A}^{k,j} \right \}:=\left \{ E_{s}^{j}(x^{k})\bigoplus E_{c}^{k}(x^{k}) | j\in \left | K \right |\right \}
\end{aligned}
\label{eq:additional_representation}
\end{equation}
These representations are extracted by combining the local invariant extractor and cross-domain specific extractors. It forces the local client to pay attention to other domains' specific attributes. These attributes may also exist in the local domain as tail attributes, which are mistakenly ignored by the local extractor. The representation augmentation improves the diversity of local distribution in latent space, mitigating the overfitting and enhancing generalization.
Besides, the final loss function of local client $k$ is replaced with:
\begin{equation}
\begin{aligned}
h_{k}&=F_{k}(\omega)+I_{s}(\omega)-I_{c}(\omega)\\
F_{k}(\omega)&=l_{CE}^{k}+\frac{\lambda }{K-1} \sum_{j=1,j\neq k}^{K}l_{CE}^{k}(R_{A}^{k,j})
\end{aligned}
\label{eq:final_loss}
\end{equation}
where the $l_{CE}$ is the cross-entropy loss; $F_{k}(\omega)$ is the loss combination of local and augmented representations; $I_{s}$, and $I_{c}$ are MI estimations. $\lambda$ are hyperparameters, which is set to 1.0 in the experiments.

{\bf Optimization process} The entire training consists of local and global two-stage alternating optimization. During each process, the parameters of some model parts are frozen for more targeted training. In the local process, the local invariant parameter is frozen for spinning off the specific attributes into the local specific branch, which is only optimized locally. Then, the optimal local specific and aggregated model parameters are frozen in the global process for better disentanglement of domain-invariant attributes. MI maximization drives the local invariant branch to concentrate on the decision-casual and cross-domain invariant attributes. Afterward, the optimal local invariant parameters are sent to the server for the next round of invariant aggregation. The diversity transfer is employed for representation augmentation in both two training stages. The detailed optimization process is in \cref{alg:expected_optimization}, and the parameter updating is shown in \cref{alg:Parameter_updating_optimization}.
\begin{algorithm}[tb]
   \caption{Optimization process of DFL.}
   \label{alg:expected_optimization}
\begin{algorithmic}
   \STATE {\bfseries Input:} The model initialization $\omega^{0}=M(\omega_{c}^{0},\omega_{s}^{0})$; The distributed Non-i.i.d datasets$\left \{ D_{k} \right \}_{k=1}^{N}$; Total optimization round $T$; Number of participating clients $K$.
   \FOR{$i=0$ {\bfseries to} $T-1$}
   \STATE The client subset of t-th round $S_{t}$ is selected from the N clients.\;
    \STATE Send the global aggregation invariant parameter $\omega_{c}^{t}$ to replace the local invariant branch.\;
   \FOR{$k\in S_{t}$}
   \STATE The optimization of the local model is divided into two steps.\;
        \STATE {\bf Step 1:} Disentanglement of local specific attributes with freezing the local invariant branch.\;
        Client k finds a $\widehat{\omega _{k}^{t+1}}=M(\omega _{c}^{t},\omega _{k,s}^{t+1,*})$ which is a $\gamma-$inexact minimizer of:\;
        \STATE $\omega _{k,s}^{t+1,*}\approx \underset{\omega _{k,s}}{\argmin}h_{k}({\omega}' ,\omega _{c}^{t})$\;
        where ${\omega}'=M(\omega_{k,c}=\omega_{c}^{t},\omega_{k,s})$\;
        
        \STATE {\bf Step 2:} Disentanglement of cross-domain invariant attributes with freezing the local specific branch.\;
        Client k finds a $\widetilde{\omega _{k}^{t+1}}=M(\widetilde{\omega _{k,c}^{t+1}},\omega _{k,s}^{t+1,*})$ which is a $\gamma-$inexact minimizer of:\;
        \STATE $\widetilde{\omega _{k,c}^{t+1}}\approx \underset{\omega _{k,c}}{\argmin}h_{k}({\omega}'' ,\omega _{c}^{t})$\;
        where ${\omega}''=M(\omega_{k,c},\omega_{k,s}=\omega_{k,s}^{t+1,*})$\;
    \ENDFOR
   \STATE {\bf Step 3:} Invariant aggregation.\;
    $\widetilde{\omega_{c}^{t+1}}=\mathbb{E}_{s_{t}}[\widetilde{\omega_{k,c}^{t+1}}]$\;
   \ENDFOR
\end{algorithmic}
\end{algorithm}
\begin{algorithm}[tb]
   \caption{Parameter updating of DFL.}
   \label{alg:Parameter_updating_optimization}
\begin{algorithmic}
   \STATE {\bf Step 1:} $\omega _{k,s}$ do not participate in model aggregation, which only update locally.
$\omega _{k,s}^{t+1}=\omega _{k,s}^{t}-\eta _{s}(P_{s}^{T}\bigtriangledown _{\omega }F_{k}(\omega )+\bigtriangledown _{\omega_{s} }I_{s}(\omega _{k,s}^{t},\omega _{c}^{t}))$\;
   \STATE {\bf Step 2:}The updating of $\omega _{k,c}^{t+1}$ based on global aggregation parameter $\omega _{c}^{t}$.
$\omega _{k,c}^{t+1}=\omega _{c}^{t}-\eta _{c}(P_{c}^{T}\bigtriangledown _{\omega }F_{k}(\omega )-\bigtriangledown _{\omega_{c} }I_{c}(\omega _{k,c}^{t},\omega _{c}^{t}))$\;
  \STATE {\bf Step 3:} invariant aggregation $\omega _{c}^{t+1}=\frac{1}{K}\underset{k\in s_{t}}{\sum} \omega _{k,c}^{t+1}$
\end{algorithmic}
\end{algorithm}

\subsection{Convergence Analysis}
\label{sec:Convergence Analysis} 
As the optimization process is alternately performed, the training process is changed from a one-stage global optimization to a two-stage partial optimization. In global optimization, only part of the extractor from each client participates in the model aggregation. This work proved the convergence of the system with the following assumptions:

{\bf 1. Non-convex and L-Lipschitz smoothness of $f$:}
    \begin{equation}
    \begin{aligned}
    \left \| \bigtriangledown f(\omega )-\bigtriangledown f(\omega' ) \right \|\leqslant L\left \|\omega -\omega'  \right \|,  \forall \omega,\omega'
    \end{aligned}
    \label{eq:Non-convex_and_L-Lipschitz_smoothness}
    \end{equation}
{\bf 2. Polyak-Łojasiewicz of $I_{c},I_{s}$:}
    \begin{equation}
    \begin{aligned}
    \left \|\bigtriangledown{I_{c}\left ( \omega ,\omega _{c}^{t} \right )}-\bigtriangledown{I_{c}\left ( \omega' ,\omega _{c}^{t} \right )}   \right \|\geqslant \mu _{I_{c}}\left \| \omega -\omega ' \right \|,  \forall \omega,\omega'\\
   \left \|\bigtriangledown{I_{s}\left ( \omega ,\omega _{c}^{t} \right )}-\bigtriangledown{I_{s}\left ( \omega' ,\omega _{c}^{t} \right )}   \right \|\geqslant \mu _{I_{s}}\left \| \omega -\omega ' \right \|,  \forall \omega,\omega'
    \end{aligned}
    \label{eq:strongly convex of Is}
    \end{equation}
{\bf 3. $\overline{\mu }-$strongly convex of $h_{k}$ and Polyak-Łojasiewicz:}
\begin{equation}
    \begin{aligned}
    &\left \| \bigtriangledown h_{k}(M(\omega _{c},\omega _{k,s}^{t+1,*}),\omega_{c}^{t})-\bigtriangledown h_{k}(M(\omega _{c}',\omega _{k,s}^{t+1,*}),\omega_{c}^{t}) \right \| \\
    &\geq \overline{\mu}  \left \| \omega _{c}-\omega _{c}' \right \|
    \end{aligned}
    \label{eq:strongly convex of Ic}
    \end{equation}
{\bf 4. Bounded second moments of $I_{c},I_{s}$ gradient:}
\begin{equation}
\begin{aligned}
\mathbb{E}_{k}\left [  \left \| \bigtriangledown I_{c}\left ( \omega ,\omega_{c}^{t} \right ) \right \|^{2}\right ]\leqslant \epsilon _{c}^{2},
\exists \epsilon _{c}\\
\mathbb{E}_{k}\left [  \left \| \bigtriangledown I_{s}\left ( \omega ,\omega_{c}^{t} \right ) \right \|^{2}\right ]\leqslant \epsilon _{s}^{2}, 
\exists \epsilon _{s}
\end{aligned}
\label{eq:Bounded second moments Is}
\end{equation}
The definitions of the $\gamma-$inexact solution and B-local dissimilarity are the same as FedProx~\cite{li2018federated} in the Appendix. First, the expected aggregation model parameter is defined as: $\overline{\omega}_{c}^{t+1}=\mathbb{E}_{k}\left [ \omega_{k,c}^{t+1} \right ]$. Then, we establish the updating relationship between the expected and empirical parameters of the aggregation model, using local-global two-stage optimization. The optimization purpose of the local and global training processes is defined in \cref{alg:expected_optimization}.
\begin{definition}
\label{def:inj}
In client $k$, local specific optimal parameter is $\widehat{\omega _{k}^{t+1}}=M(\omega _{c}^{t},\omega _{k,s}^{t+1,*})$, and the invariant optimal parameter is $\widetilde{\omega _{k}^{t+1}}=M(\widetilde{\omega _{k,c}^{t+1}},\omega _{k,s}^{t+1,*})$. Besides, the empirical parameters applied for the updating of client $k$ is $\omega _{k}^{t+1}=M(\omega _{k,c}^{t+1},\omega _{k,s}^{t+1,*})$.
\end{definition}
With the bounded gradient of local specific optimization and the empirical updating in the \cref{alg:Parameter_updating_optimization}, The following inequality can be derived:
\begin{equation}
\begin{aligned}
\left \| \widetilde{\omega _{k,c}^{t+1}}-\omega _{k,c}^{t+1} \right \| &\leqslant \frac{\gamma }{\overline{\mu }P_{c}} \left \| \bigtriangledown h_{k}(\widehat{\omega _{k}^{t+1}})\right \| \\
\left \| \omega _{k,c}^{t+1}- \omega _{c}^{t} \right \| \leqslant &  \frac{1+\gamma }{\overline{\mu }P_{c}} \left \| \bigtriangledown h_{k}(\widehat{\omega _{k}^{t+1}})\right \|
\end{aligned}
\label{eq:proof11}
\end{equation}
\cref{eq:proof11} verifies that 1) the distance between expected and empirical of the local invariant extractor's parameter; and 2) the local invariant updating are both bounded by the gradient of the local specific extractor. Thus, the updating relationship between the expected and empirical parameters is derived as:
\begin{equation}
\begin{aligned}
&\left \| \overline{\omega} _{c}^{t+1}- \omega _{c}^{t} \right \| ^{2}
\leqslant \mathbb{E}_{k}\left [\left \| \omega_{k,c}^{t+1}-\omega_{c}^{t}  \right \|^{2} \right ]\\
&\leqslant  \frac{(1+\gamma)^{2} }{\overline{\mu }^{2}P_{c}^{2}}\left (  2B^{2}\left \| \bigtriangledown f(\omega _{c}^{t}) \right \|^{2}+\left ( \frac{8L^{2}P_{s}^{2}}{\mu _{I_{s}}^{2}}+2 \right )\epsilon _{s}^{2}\right )
\end{aligned}
\end{equation}
After that, the local Lipschitz continuity of the function $f$ is applied to approximate $\omega_{c}^{t+1}$.
\begin{equation}
\begin{aligned}
& f\left ( \omega _{c}^{t+1} \right )\leqslant f\left ( \overline{\omega}_{c}^{t+1} \right )+L_{0}\left \| \omega _{c}^{t+1} - \overline{\omega}_{c}^{t+1}\right \| \\
& L_{0}\leqslant \left \| \bigtriangledown f\left ( \omega _{c}^{t} \right ) \right \|+L\left ( \left \| \overline{\omega}_{c}^{t+1} -\omega _{c}^{t} \right \|+\left \| \omega _{c}^{t+1}-\omega _{c}^{t} \right \| \right )
\end{aligned}
\label{eq:proof24}
\end{equation}
After finishing the derivation process, the convergence of DFL is obtained as:
\begin{theorem}
\label{thm:bigtheorem}
{\bf Convergence of disentangled federated learning.} Let Assumptions 1-4 hold. Suppose that $\omega_{c}^{t}$ is not a stationary solution and the local functions $F_{K}$ is B-locally dissimilar, i.e. $B(\omega_{c}^{t})\leqslant B$. If the hyperparameters in $\alpha, \beta$ are chosen such that:
\begin{equation}
\begin{aligned}
\alpha &=\frac{\eta _{c}}{2}+2\eta _{c}P_{c}^{2}B^{2}-\frac{LB^{2}(1+\gamma )^{2}}{\overline{\mu}^{2}P_{c}^{2}}\\
&-\frac{2(1+\gamma )B}{\overline{\mu}P_{c}\sqrt{K}}- \frac{2(2\sqrt{2K}+2)L(1+\gamma )^{2}B^{2}}{K\overline{\mu}^{2}P_{c}^{2}} >0\\
\beta &=\frac{\sqrt{K}\overline{\mu}P_{c}(1+\gamma )+BL(K+4\sqrt{2K}+4)(1+\gamma )^{2}}{KB\overline{\mu}^{2}P_{c}^{2}}\\
&\left ( \frac{4L^{2}P_{s}^{2}}{\mu _{I_{s}}^{2}}+1 \right )-\frac{8\eta _{c}L^{2}P_{c}^{2}P_{s}^{2}}{\mu _{I_{s}}^{2}}
\end{aligned}
\label{eq:beta}
\end{equation}
then at the t-th round of optimization, the expected decrease in the global objective is:
\begin{equation}
\begin{aligned}
\mathbb{E}_{s_{t}}[f(\omega _{c}^{t+1})]\leqslant f(\omega _{c}^{t})-\alpha \left \| \bigtriangledown f(\omega _{c}^{t}) \right \|+\beta \epsilon _{s}^{2}-\eta_{c}\epsilon _{c}^{2}
\end{aligned}
\label{eq:expected decrease}
\end{equation}
which means the convergence is that:
\begin{equation}
\begin{aligned}
\frac{1}{T}\sum_{t=0}^{T-1}\left \| \bigtriangledown f(\omega _{c}^{t}) \right \|\leqslant \frac{1}{\alpha T}\left (f(\omega _{c}^{0}) -f^{*}  \right )+\beta \epsilon _{s}^{2}-\eta_{c}\epsilon _{c}^{2}
\end{aligned}
\label{eq:convergence}
\end{equation}
where $f^{*}$ is the minimum value of the problem and the $\omega _{c}^{0}$ is the initialization of the model.  
\end{theorem}
The convergence proof is detailed in Appendix. As far as we know, \cref{thm:bigtheorem} first provides the convergence guarantee of partial aggregation. It means that the DFL system is convergent even if only part of the extractor participates in the aggregation, based on the bounded gradient of the local specific branch. Compared with FedAvg, the convergence rate of DFL can be sped up with better optimization of local specific branches. This process is only trained locally, which reduces the entire time cost with fewer communication rounds. The other benefits of this theory are that: 1) The constraint of the client's model consistency is relaxed, amplifying the FL application. 2) The incomplete model makes the raw data reconstruction more difficult, enhancing privacy security.

\begin{figure*}[t]
\begin{center}
\includegraphics[width=0.9\linewidth]{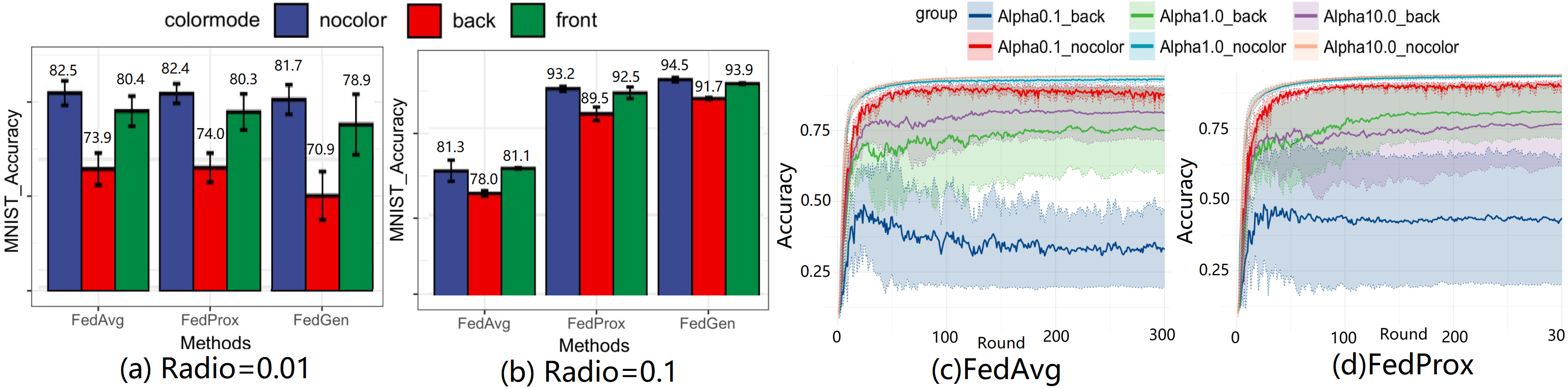}
\vskip -0.1in
\caption{(a) and (b) show the huge accuracy degradation of the MNIST classification with the introduction of attributes skew, in different FL methods and sampling ratios. Similarly, (c) and (d) represents the unstable convergence of the Colored-MNIST.}
\label{fig:representation}
\end{center}
\vskip -0.1in
\end{figure*}

\begin{table*}[t]
\begin{center}
\caption{Top-1 test accuracy of verifications on Colored-MNIST, 3Dshapes, dSprites. BG color means that each client has different floor color and wall color. }
\vskip 0.15in
\begin{tabular}{c|c|c|c|c|c|c}
\hline
Dataset & Attributes & clients & FedAvg & FedProx & FedGen & DFL \\ \hline
Colored-MNIST & BG color & 10/20 & 88.88±0.28 & 89.93±0.87 & 93.47±0.26 & \textbf{95.91±0.13} \\ \hline
3Dshapes & BG color & 20/50 & 98.57±0.46 & 98.16±0.79 & 98.38±0.47 & \textbf{99.37±0.09} \\ \hline
3Dshapes & Scale & 10/10 & 89.34±1.25 & 89.93±1.43 & 76.57±9.18 & \textbf{90.38±0.56} \\ \hline
dSprites & Orientation & 20/40 & 73.55±4.78 & 71.64±5.23 & 82.69±1.82 & \textbf{86.74±2.09} \\ \hline
\end{tabular}
\label{Tab:Verification}
\end{center}
\vskip -0.1in
\end{table*}

\begin{figure}[t]
\vskip 0.2in
\begin{center}
\includegraphics[width=0.9\linewidth]{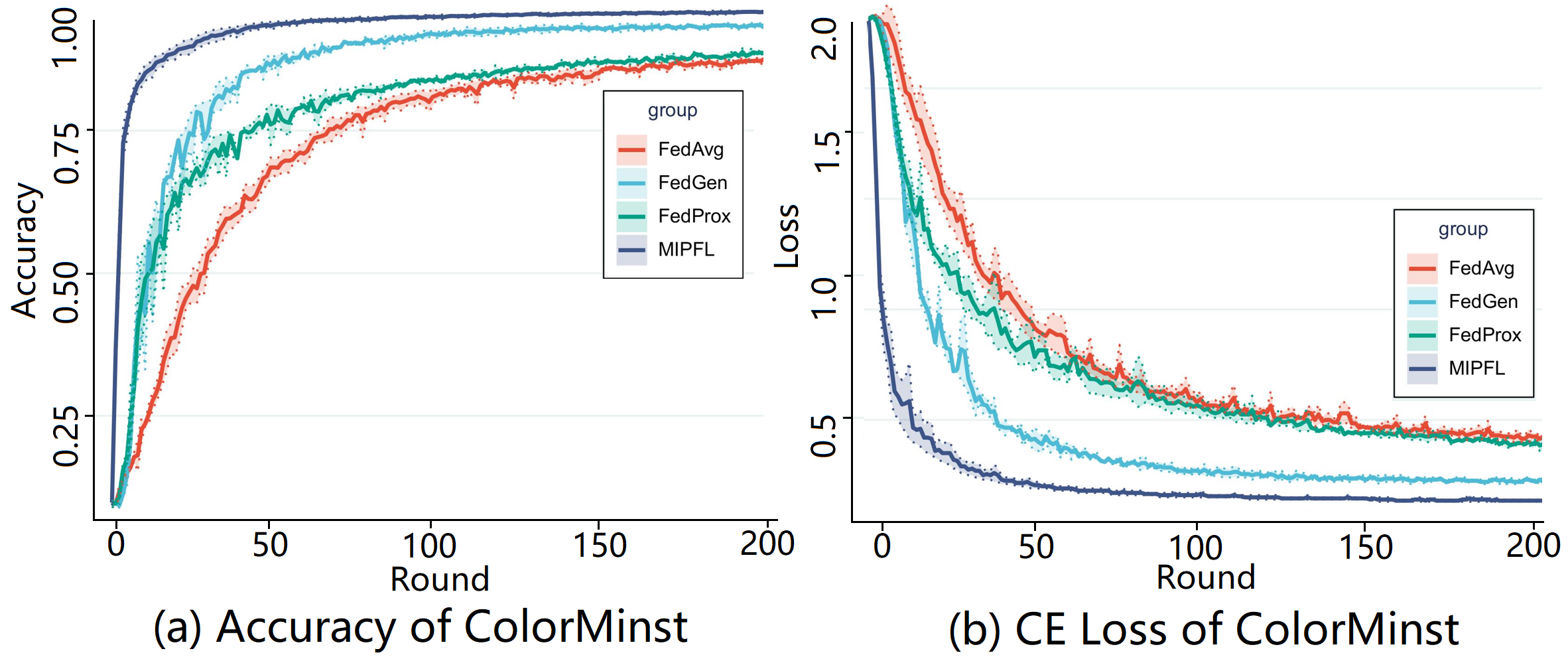}
\vskip -0.1in
\caption{Accuracy and cross-entropy loss curves with global communication rounds increase. DFL improves classification performance and convergence stability compared with other baseline FL methods in Colored-MNIST. DFL reaches better performance with fewer communication rounds.}
\label{fig:colorminst_MIFD}
\end{center}
\vskip -0.1in
\end{figure}

\begin{figure}[h]
\begin{center}
\vskip 0.2in
\includegraphics[width=0.6\linewidth]{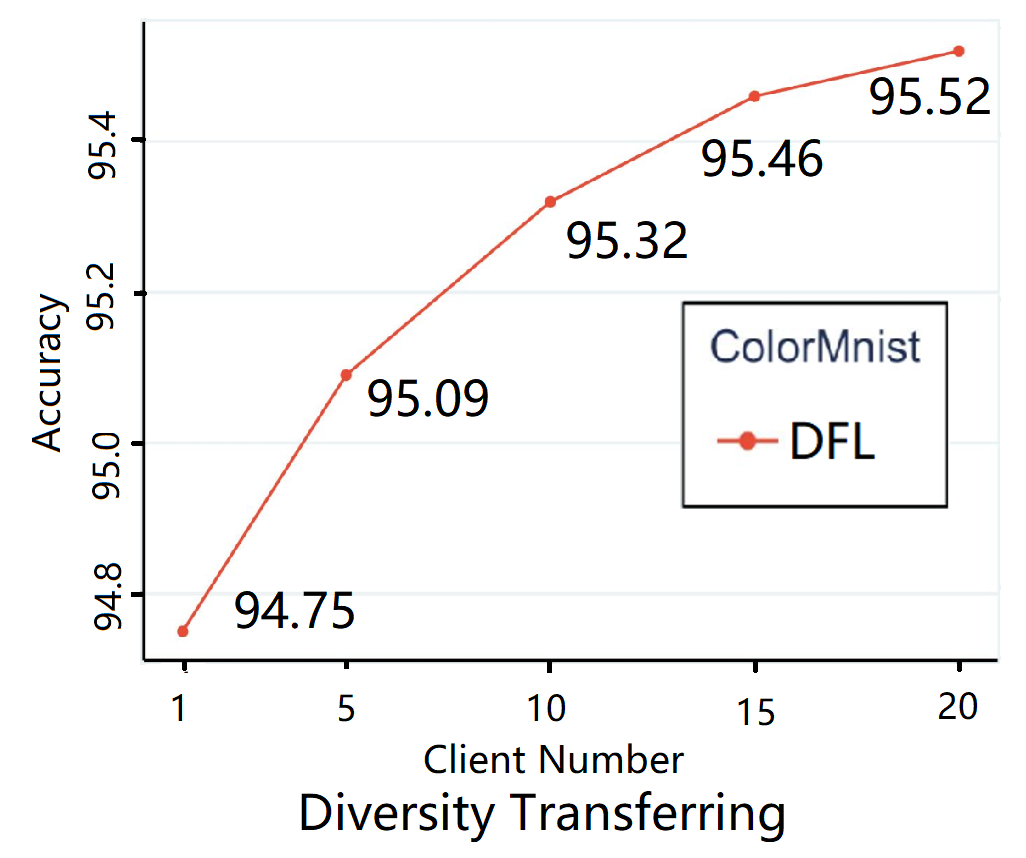}
\vskip -0.1in
\caption{Accuracy curve as client number increases.}
\label{fig:clients_number}
\end{center}
\vskip -0.3in
\end{figure}

\begin{figure*}[t]
\vskip 0.2in
\begin{center}
\includegraphics[width=0.9\linewidth]{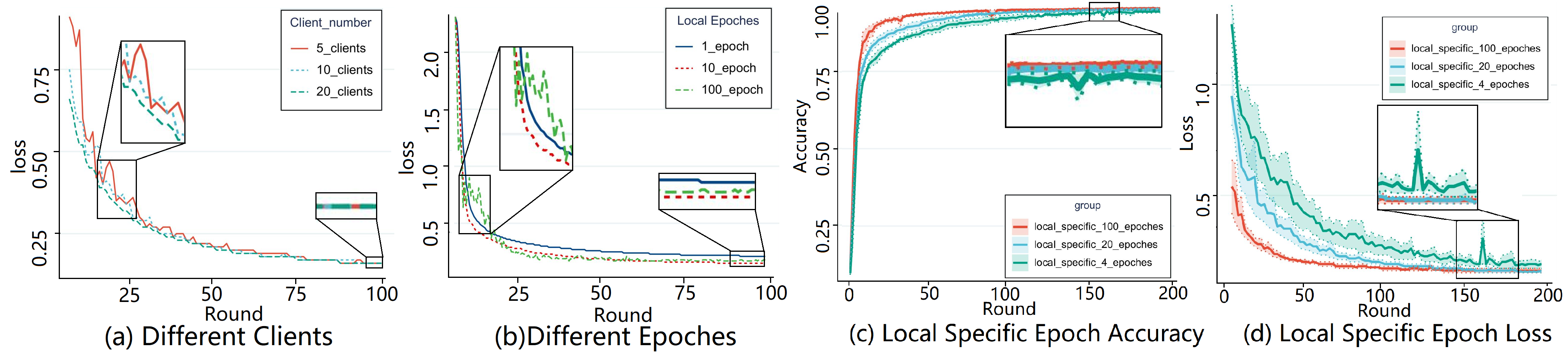}
\vskip -0.1in
\caption{(a), (b) are the loss curves with different client participation and different local updating epochs.(c), (d) are the accuracy and cross-entropy loss curves.}
\label{fig:clients_epoches}
\end{center}
\vskip -0.1in
\end{figure*}

\begin{table}[t]
\begin{center}
\caption{Ablation study of DFL in Colored-MNIST.}
\vskip 0.15in
\begin{tabular}{c|c|c|c}
\hline
           & \begin{tabular}[c]{@{}c@{}}Invariant\\ Aggregation\end{tabular} & \begin{tabular}[c]{@{}c@{}}Diversity\\ Transferring\end{tabular} & DFL               \\ \hline
10/20 & $\surd$                                      &                                                                  & 95.11±0.13          \\ \cline{2-4} 
Ratio=0.5  &                                                  & $\surd$                                      & 95.29±0.33          \\ \cline{2-4} 
BG-color   & $\surd$                                    & $\surd$                                      & \textbf{96.02±0.30} \\ \hline
\end{tabular}
\label{Tab:Ablation}
\end{center}
\vskip -0.2in
\end{table}

\begin{table}[t]
\begin{center}
\caption{Top-1 test accuracy of application on Office-Caltech10. A, C, D , and W are abbreviations for Amazon, Caltech, DSLR and WebCam.}
\vskip 0.15in
\begin{tabular}{c|ccccc}
\hline
Dataset & \multicolumn{5}{c}{Office-Caltech10, Backbone=AlexNet}                                                                                                                                                                                                             \\ \cline{2-6} 
Methods & \multicolumn{1}{c|}{A}                                & \multicolumn{1}{c|}{C}                               & \multicolumn{1}{c|}{D}                                   & \multicolumn{1}{c|}{W}                                & Avg            \\ \hline
Fedavg  & \multicolumn{1}{c|}{60.64}                                 & \multicolumn{1}{c|}{54.22}                                 & \multicolumn{1}{c|}{87.50}                                  & \multicolumn{1}{c|}{96.61}                                 & 74.69          \\ \hline
FedProx & \multicolumn{1}{c|}{59.38}                                 & \multicolumn{1}{c|}{53.33}                                 & \multicolumn{1}{c|}{84.38}                                  & \multicolumn{1}{c|}{94.92}                                 & 73.00          \\ \hline
FedBN   & \multicolumn{1}{c|}{69.27}                                 & \multicolumn{1}{c|}{{ 55.00}} & \multicolumn{1}{c|}{96.88}                                  & \multicolumn{1}{c|}{97.31}                                 & 79.61         \\ \hline
DFL   & \multicolumn{1}{c|}{ \textbf{73.96}} & \multicolumn{1}{c|}{\textbf{55.56}}                                 & \multicolumn{1}{c|}{{ \textbf{100.00}}} & \multicolumn{1}{c|}{{ \textbf{98.31}}} & \textbf{81.96} \\ \hline
\end{tabular}
\label{Tab:Verification}
\end{center}
\vskip -0.2in
\end{table}

\begin{table*}[t]
\begin{center}
\caption{Top-1 test accuracy of application on DomainNet.}
\vskip 0.15in
\begin{tabular}{cc|c|c|c|c|c|c|c}
\hline
 &  & Clipart & Infograph & Painting & Quickdraw & Real & Sketch & Avg \\ \hline
\multicolumn{1}{c|}{FedAvg} & DomainNet & 77.70 & 37.29 & 62.84 & 73.00 & 70.67 & 72.56 & 65.68 \\ \cline{1-1} \cline{3-9} 
\multicolumn{1}{c|}{FedProx} & Backbone & 77.71 & 38.96 & 62.20 & 72.50 & 71.08 & 71.12 & 65.60 \\ \cline{1-1} \cline{3-9} 
\multicolumn{1}{c|}{FedBN} & =AlexNet & 76.43 & 35.31 & 65.11 & 83.60 & 74.45 & 74.55 & 68.24 \\ \cline{1-1} \cline{3-9} 
\multicolumn{1}{c|}{DFL} & Top-10 Classes & \textbf{77.76} & \textbf{41.55} & \textbf{66.88} & \textbf{84.10} & \textbf{76.42} & \textbf{74.65} & \textbf{70.23} \\ \hline
\multicolumn{1}{c|}{FedAvg} & DomainNet & 96.32 & 60.12 & 94.83 & 82.10 & 95.81 & 93.68 & 87.14 \\ \cline{1-1} \cline{3-9} 
\multicolumn{1}{c|}{FedProx} & Backbone & 96.58 & 60.27 & 94.67 & 82.90 & 95.15 & 94.04 & 87.27 \\ \cline{1-1} \cline{3-9} 
\multicolumn{1}{c|}{FedBN} & =ResNet101 & \textbf{97.15} & 61.34 & 94.80 & 87.00 & 96.63 & 94.95 & 88.65 \\ \cline{1-1} \cline{3-9} 
\multicolumn{1}{c|}{DFL} & Top-10 Classes & 96.20 & \textbf{61.64} & \textbf{95.01} & \textbf{89.60} & \textbf{96.73} & \textbf{95.67} & \textbf{89.14} \\ \hline
\multicolumn{1}{c|}{SingleSet} & ResNet101 & 69.3 & 34.5 & 66.3 & 66.8 & 80.1 & 60.7 & 62.95 \\ \cline{1-1} \cline{3-9} 
\multicolumn{1}{c|}{DFL} & All 345 Classes & \textbf{78.4} & \textbf{38.2} & \textbf{71.2} & \textbf{70.4} & \textbf{82.7} & \textbf{68.6} & \textbf{68.25} \\ \hline
\end{tabular}
\label{Tab:Verification_domainnet}
\end{center}
\vskip -0.2in
\end{table*}

\section{Experiment}
\label{sec:experiment}
This section consists of three experimental parts: clarification, verification, and application. The first part shows the complex problems caused by attributes skew in FL, which is the independent component of Non-i.i.d and exists ubiquitously. Clarification experiments clarify that attributes skew can cause unstable convergence and performance degradation. Verification experiments focus on manually synthesized attributes skew, which aims to verify the effectiveness, loss convergence, and performance improvement of DFL compared with other critical related works. Application experiments pay attention to the performance of DFL in datasets with realistic attributes skew, which tries to prove that DFL can adapt to the practical environment. 

{\bf Benchmark datasets}: The clarification experiments are performed on the MNIST and {\bf colored-MNIST~\cite{arjovsky2019invariant}}. The former is attributes balanced, and the latter has skewed attributes as different foreground/background colors in different clients. These digit classification datasets are both divided into 20 clients. The verification experiments are performed on: 1) {\bf colored-MNIST}~\cite{arjovsky2019invariant} with attributes skew as the background color (BG color in \cref{Tab:Verification}). 2) {\bf 3dshapes}~\cite{3dshapes18} is employed the background colors (BG color in \cref{Tab:Verification}) or scales(Scale in \cref{Tab:Verification}) as attributes skew. The shape is employed for classification. 3) {\bf dSprites}~\cite{dsprites17} performs object scale classification, with the attributes skew as orientation (Orientation in \cref{Tab:Verification}). Besides, the training sampling ratio in verification experiments is reduced to increase the classification difficulty. However, one thing is sure the comparisons are fair for every FL method in each experiment. 
The application experiments employ two datasets: 1) {\bf Office-Caltech10}~\cite{gong2012geodesic}, which contains ten overlapping classes from four domains acquired in different cameras or environments. 2) {\bf DomainNet}~\cite{peng2019moment}  has 345 overlapping classes from six domains with different image styles. For both application datasets, the representation distribution across attributes on each domain is different from each other. In addition, both whole training datasets are employed for distributed training.

{\bf Backbones:} For clarification and verification, the network architecture follows as ~\cite{mcmahan2017communication}. The representation extraction networks contain three groups of Conv-BN-Relu layers. The classifier is composed of two fully connected layers. For application, the AlexNet~\cite{krizhevsky2012imagenet} without pre-training and ResNet101~\cite{he2016deep} with pre-training are selected as representation extraction modules. The classifier consists of 3 fully connected layers and two batch normalization layers. 

{\bf Comparison}  Five pioneering FL methods are fairly compared. FedAvg~\cite{mcmahan2017communication} is the classic FL using simple aggregation. FedProx~\cite{li2018federated} employed regularization terms for optimization correction that achieve good personalized performance. FedGen~\cite{zhu2021data} proposed a data-free knowledge distillation to mitigate the distribution shift. FedEnsemble~\cite{shi2021fed} introduced model ensembling to FL using random permutations for updating. FedBN~\cite{li2021fedbn} applied local batch normalization to alleviate the feature shift.

{\bf Configurations:} Unless otherwise mentioned, the global communication round set 200 for clarification/verification and 100 for application. The testing sample ratio of verification experiments is 0.5, which means half of the images are used for testing. The training sample ratios are set to: 1) 0.01 or 0.1 for MNIST and Colored-MNIST in the clarification experiments; 2) 0.5 for Colored-MNIST, 0.01 for 3Dshapes, and 0.2 for dSprites in verification experiments; 3) 1.0 for Office-Caltech10 and DomainNet in the application experiments, which means using all training data for FL training. The data distribution of the training and testing set is the same in each experiment in one client, but different between clients. The local updating step is E=20, and the mini-batch size is B=32. The learning rate lr=0.0.1. The total client numbers are 20 for Colored-MNIST, 50 for 3Dshapes with BG color skew, 10 for 3Dshapes with scale skew, 40 for dSprites, 4 for Office-Caltech10, and 6 for DomainNet.

\subsection{Clarification}
(a) and (b) of \cref{fig:representation} shows huge accuracy degradation with the introduction of attributes skew. With the decrease in sampling ratio, the performance damage rapid growth, which increased from less than 4\% at 0.1 ratio to over 10\% at 0.01 ratio. In addition, the negative transfer caused by the background color attributes skew is worse than the foreground color skew. To further explore the influence of attributes skew on the FL system,  the attributes skew and label skew are mixed, which is closer to reality. The accuracy curves as (c) and (d) in \cref{fig:representation} show both the performance and convergence are severely damaged in any FL methods, which even tends to be un-convergent. In summary, the stability, convergence, and performance of the FL system suffer from the introduction of attributes skew. Worse yet, this phenomenon widely exists in reality, which seriously hinders the application of FL. Thus, It is crucial to delve into and mitigate the bad influence of attributes skew.

\subsection{Verification} 
{\bf Performance:} The verification results are shown in \cref{Tab:Verification}. DFL outperforms FedAvg, FedProx, and FedGen on all three datasets with different attributes skew and different classification tasks. Especially, FedGen applied global representation generation to complement local latent space. The higher performance of DFL means that the diversity transferring augmented the local representations with a more accurate distribution. The accuracy curve as (a) in \cref{fig:colorminst_MIFD}  indicates that DFL provides a faster and more stable performance improvement. 

{\bf Convergence:} The cross-entropy loss curve as (b) in \cref{fig:colorminst_MIFD} shows that DFL achieves a higher rate and more stable convergence compared with SOTA FL methods. The small volatility of the DFL loss curve verifies the similar optimization directions provided by invariant aggregation. The accuracy and loss curve of (c) and (d) in \cref{fig:clients_epoches} represent that the convergence is sped up, and the accuracy is improved with the increase of local specific training epochs. It also means that the DFL can achieve better performance with fewer communication rounds. In addition, the influence of participating clients number increasing is analyzed. The loss curve as (a) in \cref{fig:clients_epoches} shows that the convergence is more stable with more clients participating, although the final losses are the same. Similarly, ten local updating epochs setting achieves the best and fastest convergence compared with 1 or 100, shown in (b) of \cref{fig:clients_epoches}. 

{\bf Communication cost:} At first, the (c) and (d) of \cref{fig:clients_epoches} present that the convergence rate is sped up due to better local specific optimization, which is trained independently with more local epochs. It means DFL needs fewer communication rounds. Second, diversity transferring is an optional component. Analyzing the results in Table 1 and the ablation study in Table 2, DFL only with invariant aggregation has fewer communication costs and outperforms other SOTA FL methods. Third, \cref{fig:clients_number} quantitatively presents the improvement of absolute accuracy as client number increases, wherein more clients participating in the diversity transferring means more communication cost. The performance gain tends to stabilize with more participating clients. It means that only a few clients selected are enough for diversity transferring. The communication overhead will not be very large.

{\bf Ablation:} \cref{Tab:Ablation} is the ablation study of DFL, which verifies the complementarity of invariant aggregation and diversity transferring. The one-stage optimization is also tested, but the training diverged, and the results are not shown in the body text.

\begin{figure}[t]
\begin{center}
\includegraphics[width=1.0\linewidth]{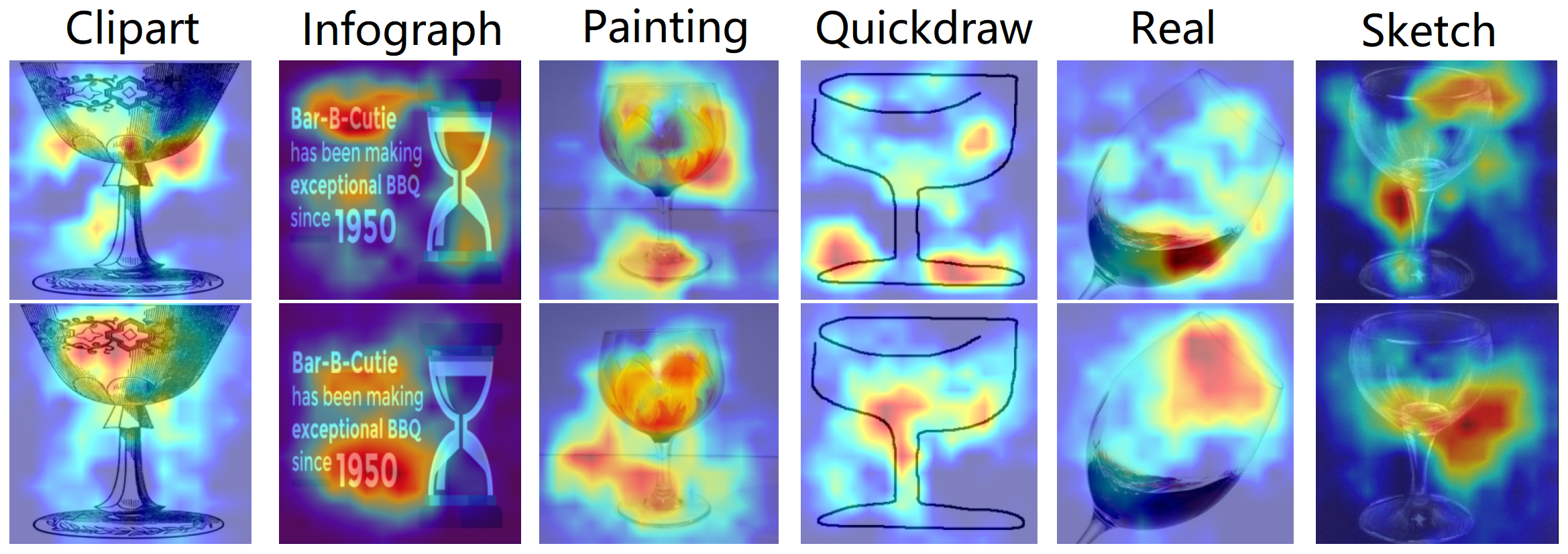}
\vskip -0.1in
\caption{Visualization of DomainNet. The first line shows the Grad-CAM~\cite{selvaraju2017grad} generated by invariant branches, and the second line shows specific branches.}
\label{fig:visualization}
\end{center}
\vskip -0.1in
\end{figure}

\subsection{Application}
Application experiments verify that DFL can adapt to realistic environments with realistic attribute shifts. \cref{Tab:Verification} and \cref{Tab:Verification_domainnet} show that  DFL outperforms other baselines with a considerable margin.  In Office-Caltech10, DFL significantly improves all categories and mean accuracy by almost 3\%. For DomainNet, regardless of whether the backbone is AlexNet or ResNet101, the mean accuracy of DFL is improved. In addition, the performance of DFL is better than SingleSet training, which is only trained by the source data. It proves that DFL transfers knowledge from other domains successfully. The visualization in \cref{fig:visualization} is an example of mug classification from DomainNet. The heatmaps of invariant branches are highlighted around the shape of the objects, which is causal to the final decision. The specific branches from different clients pay attention to different attributes such as liquid, description, and color, which are decision-correlated but domain-specific. It demonstrates the effectiveness of disentanglement in DFL. In future work, we would implement DFL on MindSpore\footnote{https://www.mindspore.cn/}, which is a new deep learning computing framework suitable for FL applications.

\section{Conclusion}
In this paper, we propose a novel FL paradigm that applies MI-based disentanglement to mitigate the negative transfer caused by distributed attributes skew. The invariant aggregation is proposed to spin off the mixed domain-specific attributes, which essentially corrects the optimization direction. The proposed diversity transferring augmented the representation in local latent space. The convergence of global aggregation using an incomplete client model expands the application scope of FL. Extensive experiments, guided by our solid convergence analysis, verify that DFL benefits FL with higher performance, better interpretability, and fewer communication rounds.

\section*{Acknowledgement}
This work is supported by National Natural Science Foundation of China (Grant No.62006225, 61906199, 62071468), the Strategic Priority Research Program of Chinese Academy of Sciences (CAS) (Grant No. XDA27040700), and sponsored by CAAI-Huawei Mindspore Open Fund. 
\nocite{langley00}

\bibliography{example_paper}
\bibliographystyle{icml2022}

\newpage
\appendix
\onecolumn
\section{Appendix A}
\subsection{Proof of Convergence}

{\bf Theorem 3.2. Convergence of disentangled federated learning.} Let Assumptions 1-4 hold. Suppose that $\omega_{c}^{t}$ is not a stationary solution and the local functions $F_{K}$ is B-locally dissimilar, i.e. $B(\omega_{c}^{t})\leqslant B$. If the hyperparameters in $\alpha, \beta$ are chosen such that:
\begin{equation}
\begin{aligned}
\alpha =\frac{\eta _{c}}{2}+2\eta _{c}P_{c}^{2}B^{2}-\frac{LB^{2}(1+\gamma )^{2}}{\overline{\mu}^{2}P_{c}^{2}}
-\frac{2(1+\gamma )B}{\overline{\mu}P_{c}\sqrt{K}}- \frac{2(2\sqrt{2K}+2)L(1+\gamma )^{2}B^{2}}{K\overline{\mu}^{2}P_{c}^{2}} >0\\
\beta =\frac{\sqrt{K}\overline{\mu}P_{c}(1+\gamma )+BL(K+4\sqrt{2K}+4)(1+\gamma )^{2}}{KB\overline{\mu}^{2}P_{c}^{2}}
\left ( \frac{4L^{2}P_{s}^{2}}{\mu _{I_{s}}^{2}}+1 \right )-\frac{8\eta _{c}L^{2}P_{c}^{2}P_{s}^{2}}{\mu _{I_{s}}^{2}}
\end{aligned}
\label{eq:beta}
\end{equation}
then at the t-th round of optimization, the expected decrease in the global objective is:
\begin{equation}
\begin{aligned}
\mathbb{E}_{s_{t}}[f(\omega _{c}^{t+1})]\leqslant f(\omega _{c}^{t})-\alpha \left \| \bigtriangledown f(\omega _{c}^{t}) \right \|+\beta \epsilon _{s}^{2}-\eta_{c}\epsilon _{c}^{2}
\end{aligned}
\label{eq:expected decrease}
\end{equation}
which means the convergence is that:
\begin{equation}
\begin{aligned}
\frac{1}{T}\sum_{t=0}^{T-1}\left \| \bigtriangledown f(\omega _{c}^{t}) \right \|\leqslant \frac{1}{\alpha T}\left (f(\omega _{c}^{0}) -f^{*}  \right )+\beta \epsilon _{s}^{2}-\eta_{c}\epsilon _{c}^{2}
\end{aligned}
\label{eq:convergence}
\end{equation}
where $f^{*}$ is the minimum value of the problem, and the $\omega _{c}^{0}$ is the initialization of the model.  

\begin{proof}
\begin{definition}
\label{def:γ−inexact}
{\bf $\gamma-$inexact solution:} For local function: $ h_{k}(M(\omega_{c},\omega_{s}) ,\omega _{c}^{t}) = F_{k}(M(\omega_{c},\omega_{s}))+I_{s}(\omega_{s},\omega_{c}^{t})-I_{c}(\omega_{c},\omega_{c}^{t})$, and $\gamma \in [0,1]$, $\omega_{k,c}^{*}$ is defined as $\gamma-$inexact solution of $\underset{\omega_{k,c}}{min}\left \{  h_{k}(M(\omega_{k,c},\omega_{k,s}^{t+1,*}) ,\omega _{c}^{t})  \right \}$ if:
\begin{equation}
\begin{aligned}
\left \| \bigtriangledown h_{k}(M(\omega_{k,c}^{*},\omega_{k,s}^{t+1,*}) ,\omega _{c}^{t})  \right \| \leqslant \gamma \left \| \bigtriangledown h_{k}(M(\omega_{c}^{t},\omega_{k,s}^{t+1,*}) ,\omega _{c}^{t})  \right \|
\end{aligned}
\label{eq:inexact_solution}
\end{equation}
\end{definition}

\begin{definition}
\label{def:B-local dissimilarity}
{\bf  B-local dissimilarity}: The local functions $F_{k}$ is B-locally dissimilar at $\omega_{c}^{t}$ if:
\begin{equation}
\begin{aligned}
\mathbb{E}_{k}[\left \| \bigtriangledown F_{k}(M(\omega _{c}^{t},\omega _{k,s}^{*})) \right \|^{2}] \leqslant \left \| \bigtriangledown f(\omega _{c}^{t}) \right \|^{2} B^{2}
\end{aligned}
\label{eq:B-local dissimilarity}
\end{equation}
and the $B$ is defined as:
\begin{equation}
\begin{aligned}
B(\omega _{c}^{t})=\sqrt{\frac{\mathbb{E}_{k}[\left \| \bigtriangledown F_{k}(M(\omega _{c}^{t},\omega _{k,s}^{*})) \right \|^{2}]}{\left \| \bigtriangledown f(\omega _{c}^{t}) \right \|^{2}}}
\end{aligned}
\label{eq:B}
\end{equation}
\end{definition}

\begin{definition}
\label{def:inj}
 In global, the expected aggregation model parameters is $\overline{\omega}_{c}^{t+1}=\mathbb{E}_{k}\left [ \omega_{k,c}^{t+1} \right ]$. In client $k$, local specific optimal parameter is $\widehat{\omega _{k}^{t+1}}=M(\omega _{c}^{t},\omega _{k,s}^{t+1,*})$, and the invariant optimal parameter is $\widetilde{\omega _{k}^{t+1}}=M(\widetilde{\omega _{k,c}^{t+1}},\omega _{k,s}^{t+1,*})$. Besides, the empirical parameters applied for the updating of client $k$ is $\omega _{k}^{t+1}=M(\omega _{k,c}^{t+1},\omega _{k,s}^{t+1,*})$.
\end{definition}

Let Assumptions 1-4 hold, we have:
\begin{equation}
\begin{aligned}
\left \| \overline{\omega}_{c}^{t+1}-\omega_{c}^{t} \right \|\leqslant \mathbb{E}_{k}\left [\left \| \omega_{k,c}^{t+1}-\omega_{c}^{t}  \right \| \right ]
\end{aligned}
\label{eq:proof2}
\end{equation}

\begin{equation}
\begin{aligned}
\left \|\widehat{\omega _{k}^{t+1}}-\widetilde{\omega _{k}^{t+1}}  \right \|=P_{c}\left \|\widetilde{\omega _{k,c}^{t+1}}- \omega _{c}^{t} \right \|
\end{aligned}
\label{eq:proof5}
\end{equation}

\begin{equation}
\begin{aligned}
\left \|\widehat{\omega _{k}^{t+1}}-\widetilde{\omega _{k}^{t+1}}  \right \| \leqslant \frac{1}{\overline{\mu }}\left \| \bigtriangledown h_{k}(\widehat{\omega _{k}^{t+1}})-\bigtriangledown h_{k}(\widetilde{\omega _{k}^{t+1}}) \right \|&=\frac{1}{\overline{\mu }}\left \| \bigtriangledown h_{k}(\widehat{\omega _{k}^{t+1}})\right \|
\end{aligned}
\label{eq:proof6}
\end{equation}
Thus,
\begin{equation}
\begin{aligned}
\left \|\widetilde{\omega _{k,c}^{t+1}}- \omega _{c}^{t} \right \| \leqslant  \frac{1}{\overline{\mu }P_{c}}\left \| \bigtriangledown h_{k}(\widehat{\omega _{k}^{t+1}})\right \|
\end{aligned}
\label{eq:proof7}
\end{equation}
Similar,
\begin{equation}
\begin{aligned}
\left \|\widetilde{\omega _{k}^{t+1}}- \omega _{k}^{t+1} \right \| =P_{c}\left \| \widetilde{\omega _{k,c}^{t+1}}-\omega _{k,c}^{t+1} \right \|
\leqslant \frac{1}{\overline{\mu }} \left \|\bigtriangledown h_{k}(\omega _{k}^{t+1} ) \right \|
\leqslant \frac{\gamma }{\overline{\mu }} \left \| \bigtriangledown h_{k}(\widehat{\omega _{k}^{t+1}})\right \|
\end{aligned}
\label{eq:proof8}
\end{equation}

\begin{equation}
\begin{aligned}
\left \| \widetilde{\omega _{k,c}^{t+1}}-\omega _{k,c}^{t+1} \right \| \leqslant \frac{\gamma }{\overline{\mu }P_{c}} \left \| \bigtriangledown h_{k}(\widehat{\omega _{k}^{t+1}})\right \|
\end{aligned}
\label{eq:proof9}
\end{equation}

Trigonometric inequality is introduced:
\begin{equation}
\begin{aligned}
\left \| \omega _{k,c}^{t+1}- \omega _{c}^{t} \right \|\leqslant \left \|\widetilde{\omega _{k,c}^{t+1}}- \omega _{c}^{t} \right \|+\left \| \widetilde{\omega _{k,c}^{t+1}}-\omega _{k,c}^{t+1} \right \|
\end{aligned}
\label{eq:proof10}
\end{equation}
Thus,
\begin{equation}
\begin{aligned}
\left \| \omega _{k,c}^{t+1}- \omega _{c}^{t} \right \|\leqslant  \frac{1+\gamma }{\overline{\mu }P_{c}} \left \| \bigtriangledown h_{k}(\widehat{\omega _{k}^{t+1}})\right \|
\end{aligned}
\label{eq:proof11}
\end{equation}

\begin{equation}
\begin{aligned}
\left \| \overline{\omega} _{c}^{t+1}- \omega _{c}^{t} \right \| ^{2}  \leqslant  \mathbb{E}_{k}\left [ \left \| \omega _{k,c}^{t+1}- \omega _{c}^{t} \right \| ^{2} \right ] 
\leqslant \frac{(1+\gamma)^{2} }{\overline{\mu }^{2}P_{c}^{2}}\mathbb{E}_{k}\left [ \left \| \bigtriangledown h_{k}(\widehat{\omega _{k}^{t+1}})\right \|^{2} \right ]
\end{aligned}
\label{eq:proof12}
\end{equation}

Next, the expectation calculation is introduced: 
\begin{equation}
\begin{aligned}
&\mathbb{E}_{k}\left [ \left \| \bigtriangledown h_{k}(\widehat{\omega _{k}^{t+1}})\right \|^{2} \right ]
=\mathbb{E}_{k}\left [ \left \|  \bigtriangledown  F_{k}\left ( M(\omega _{c}^{t},\omega _{k,s}^{t+1,*}) \right ) +\bigtriangledown I_{s}\left ( \omega _{k,s}^{t+1,*}, \omega _{c}^{t}\right ) \right \| ^{2} \right ]\\
&\leqslant 2\mathbb{E}_{k}\left [ \left \| \bigtriangledown  F_{k}\left ( M(\omega _{c}^{t},\omega _{k,s}^{t+1,*}) \right ) \right \| ^{2} \right ]
+2\mathbb{E}_{k}\left [ \left \| \bigtriangledown I_{s}\left ( \omega _{k,s}^{t+1,*}, \omega _{c}^{t}\right ) \right \| ^{2} \right ]\\
&\leqslant  2\mathbb{E}_{k}\left [ \left \| \bigtriangledown  F_{k}\left ( M(\omega _{c}^{t},\omega _{k,s}^{t+1,*}) \right ) \right \| ^{2} \right ]+2\epsilon _{s}^{2}
\end{aligned}
\label{eq:proof13}
\end{equation}

\begin{equation}
\begin{aligned}
\left \| \bigtriangledown  F_{k}\left ( M(\omega _{c}^{t},\omega _{k,s}^{t+1,*}) \right ) \right \| ^{2} 
&\leqslant 2\left \| \bigtriangledown  F_{k}\left ( M(\omega _{c}^{t},\omega _{k,s}^{*}) \right ) \right \|^{2}  
 +  2\left \| \bigtriangledown  F_{k}\left ( M(\omega _{c}^{t},\omega _{k,s}^{t+1,*}) \right ) -\bigtriangledown  F_{k}\left ( M(\omega _{c}^{t},\omega _{k,s}^{*}) \right ) \right \|^{2}\\
&\leqslant 2\left \| \bigtriangledown  F_{k}\left ( M(\omega _{c}^{t},\omega _{k,s}^{*}) \right ) \right \|^{2}
+2L^{2}P_{s}^{2}\left \|\omega _{k,s}^{t+1,*}-\omega _{k,s}^{*}  \right \|^{2}
\end{aligned}
\label{eq:proof14}
\end{equation}

\begin{equation}
\begin{aligned}
\left \|\omega _{k,s}^{t+1,*}-\omega _{k,s}^{*}  \right \|^{2} 
\leqslant \frac{1}{\mu _{I_{s}}^{2}}\left \| \bigtriangledown I_{s}\left ( \omega _{k,s}^{t+1,*},\omega _{c}^{t} \right )-\bigtriangledown I_{s}\left ( \omega _{k,s}^{*},\omega _{c}^{t} \right ) \right \|^{2}
\leqslant \frac{4}{\mu _{I_{s}}^{2}}\left \| \bigtriangledown I_{s}\left ( \omega _{k,s}^{*},\omega _{c}^{t} \right ) \right \|^{2} 
\leqslant\frac{4}{\mu _{I_{s}}^{2}}\epsilon _{s}^{2}
\end{aligned}
\label{eq:proof15}
\end{equation}

\begin{equation}
\begin{aligned}
\mathbb{E}_{k}\left [ \left \| \bigtriangledown  F_{k}\left ( M(\omega _{c}^{t},\omega _{k,s}^{*}) \right ) \right \|^{2}\right ]\leqslant B^{2}\left \| \bigtriangledown f(\omega _{c}^{t}) \right \|^{2}
\end{aligned}
\label{eq:proof16}
\end{equation}

\begin{equation}
\begin{aligned}
\mathbb{E}_{k}\left [ \left \| \bigtriangledown h_{k}(\widehat{\omega _{k}^{t+1}})\right \|^{2} \right ]
\leqslant 2B^{2}\left \| \bigtriangledown f(\omega _{c}^{t}) \right \|^{2}+\left ( \frac{8L^{2}P_{s}^{2}}{\mu _{I_{s}}^{2}}+2 \right )\epsilon _{s}^{2}
\end{aligned}
\label{eq:proof17}
\end{equation}

\begin{equation}
\begin{aligned}
\left \| \overline{\omega} _{c}^{t+1}- \omega _{c}^{t} \right \| ^{2}
\leqslant \mathbb{E}_{k}\left [\left \| \omega_{k,c}^{t+1}-\omega_{c}^{t}  \right \|^{2} \right ]
\leqslant  \frac{(1+\gamma)^{2} }{\overline{\mu }^{2}P_{c}^{2}}\left (  2B^{2}\left \| \bigtriangledown f(\omega _{c}^{t}) \right \|^{2}+\left ( \frac{8L^{2}P_{s}^{2}}{\mu _{I_{s}}^{2}}+2 \right )\epsilon _{s}^{2}\right )
\end{aligned}
\label{eq:proof18}
\end{equation}

Based on the L-Lipschitz smoothness of f and Taylor expansion, it is:
\begin{equation}
\begin{aligned}
f\left ( \overline{\omega} _{c}^{t+1} \right ) &\leqslant f\left (\omega _{c}^{t} \right )+\left \langle \bigtriangledown f\left (\omega _{c}^{t} \right ), \overline{\omega} _{c}^{t+1}-\omega _{c}^{t} \right \rangle
+\frac{L}{2}\left \| \overline{\omega} _{c}^{t+1}-\omega _{c}^{t} \right \|^{2}
\end{aligned}
\label{eq:proof19}
\end{equation}

\begin{equation}
\begin{aligned}
\mathbb{E}_{s_{t}}\left [ \left \langle \bigtriangledown f\left (\omega _{c}^{t} \right ), \overline{\omega} _{c}^{t+1}-\omega _{c}^{t} \right \rangle \right ]
=\mathbb{E}_{s_{t}}\left [ \left \langle \bigtriangledown f\left (\omega _{c}^{t} \right ), \mathbb{E}_{k}[\omega _{k,c}^{t+1}-\omega _{c}^{t}] \right \rangle \right ]
=-\eta _{c}\mathbb{E}_{s_{t}}\left [ \left \langle \bigtriangledown f\left (\omega _{c}^{t} \right ), \mathbb{E}_{k}[\bigtriangledown _{\omega _{c}}h_{k}(\widehat{\omega_{k}^{t+1}})] \right \rangle \right ]
\end{aligned}
\label{eq:proof20}
\end{equation}

Since the following inequality established as:
\begin{equation}
\begin{aligned}
\left \langle a,b \right \rangle=\frac{1}{2}\left ( \left \| a \right \|^{2}+\left \| b \right \|^{2}-\left \| a-b \right \|^{2} \right )
\end{aligned}
\label{eq:proof21}
\end{equation}
similarly, we have:
\begin{equation}
\begin{aligned}
\mathbb{E}_{s_{t}}\left [ \left \langle \bigtriangledown f\left (\omega _{c}^{t} \right ), \overline{\omega} _{c}^{t+1}-\omega _{c}^{t} \right \rangle \right ]
&\leqslant -\frac{\eta _{c}}{2}\mathbb{E}_{s_{t}}( \left \| \bigtriangledown f\left (\omega _{c}^{t} \right )  \right \|^{2}+\left \| \bigtriangledown _{\omega _{c}}h_{k}(\widehat{\omega_{k}^{t+1}}) \right \|^{2}
-\left \|\bigtriangledown f\left (\omega _{c}^{t} \right )-\mathbb{E}_{k}[\bigtriangledown _{\omega _{c}}h_{k}(\widehat{\omega_{k}^{t+1}})]  \right \|^{2} )\\
&\leqslant -\frac{\eta _{c}}{2}(\left \| \bigtriangledown f\left (\omega _{c}^{t} \right )  \right \|^{2}
+\mathbb{E}_{s_{t}}\left [ \left \| \bigtriangledown _{\omega _{c}}h_{k}(\widehat{\omega_{k}^{t+1}}) \right \|^{2} \right ])\\ 
\end{aligned}
\label{eq:proof22}
\end{equation}

\begin{equation}
\begin{aligned}
 \mathbb{E}_{s_{t}}\left [ \left \| \bigtriangledown _{\omega _{c}}h_{k}(\widehat{\omega_{k}^{t+1}}) \right \|^{2} \right ]
 &=\mathbb{E}_{s_{t}}\left [ \left \|P_{c}\bigtriangledown F_{k}\left (  \widehat{\omega_{k}^{t+1}}\right ) -\bigtriangledown I_{c}\left ( \omega _{k,c}^{t},\omega _{c}^{t}, \right ) \right \|^{2} \right ]
 \leqslant 2P_{c}^{2}\mathbb{E}_{s_{t}}\left [ \left \| \bigtriangledown F_{k}\left (  \widehat{\omega_{k}^{t+1}}\right ) \right \|^{2}\right ]+2\epsilon _{c}^{2}\\
& \leqslant 4P_{c}^{2}B^{2}\left \| \bigtriangledown f(\omega_{c}^{t}) \right \|^{2}+\frac{16L^{2}P_{c}^{2}P_{s}^{2}}{\mu _{I_{s}}^{2}}\epsilon _{s}^{2}+2\epsilon _{c}^{2}
\end{aligned}
\label{eq:proof23}
\end{equation}

The local Lipschitz continuity of the function $f$ is applied to approximate $\omega_{c}^{t+1}$.
\begin{equation}
\begin{aligned}
& f\left ( \omega _{c}^{t+1} \right )\leqslant f\left ( \overline{\omega}_{c}^{t+1} \right )+L_{0}\left \| \omega _{c}^{t+1} - \overline{\omega}_{c}^{t+1}\right \| \\
& L_{0}\leqslant \left \| \bigtriangledown f\left ( \omega _{c}^{t} \right ) \right \|+L\left ( \left \| \overline{\omega}_{c}^{t+1} -\omega _{c}^{t} \right \|+\left \| \omega _{c}^{t+1}-\omega _{c}^{t} \right \| \right )
\end{aligned}
\label{eq:proof24}
\end{equation}
Following the derivation process, we have:

\begin{equation}
\begin{aligned}
\mathbb{E}_{s_{t}}\left [ f(\omega _{c}^{t+1}) \right ]
&\leqslant f(\overline{\omega}_{c}^{t+1})
+\sqrt{\frac{2}{K}\mathbb{E}_{k}\left [ \left \|  \omega_{k,c}^{t+1}-\omega_{c}^{t} \right \|^{2} \right ]}\left \| \bigtriangledown f(\omega _{c}^{t}) \right \|
+\frac{(2\sqrt{2K}+2)L}{K}\mathbb{E}_{k}\left [ \left \|  \omega_{k,c}^{t+1}-\omega_{c}^{t} \right \|^{2}\right ]\\
& \leqslant  f(\overline{\omega}_{c}^{t+1})
+\frac{2(1+\gamma )B}{\overline{\mu}P_{c}\sqrt{K}}\sqrt{\left \| \bigtriangledown f(\omega _{c}^{t}) \right \|^{4}+\left ( \frac{4L^{2}P_{s}^{2}}{\mu _{I_{s}}^{2}}+1 \right )\frac{\epsilon _{s}^{2}}{B^{2}}\left \| \bigtriangledown f(\omega _{c}^{t}) \right \|^{2}}\\
& +\frac{2(2\sqrt{2K}+2)L(1+\gamma )^{2}}{K\overline{\mu}^{2}P_{c}^{2}}\left ( B^{2} \left \| \bigtriangledown f(\omega _{c}^{t}) \right \|^{2}+\left ( \frac{4L^{2}P_{s}^{2}}{\mu _{I_{s}}^{2}}+1 \right )\epsilon _{s}^{2}\right )\\
&\leqslant f(\overline{\omega}_{c}^{t+1})
+\left ( \frac{2(1+\gamma )B}{\overline{\mu}P_{c}\sqrt{K}}+ \frac{2(2\sqrt{2K}+2)L(1+\gamma )^{2}B^{2}}{K\overline{\mu}^{2}P_{c}^{2}}\right )\left \| \bigtriangledown f(\omega _{c}^{t}) \right \|^{2}\\
&+\frac{\sqrt{K}\overline{\mu}P_{c}(1+\gamma )+2BL(2\sqrt{2K}+2)(1+\gamma )^{2}}{KB\overline{\mu}^{2}P_{c}^{2}}\left ( \frac{4L^{2}P_{s}^{2}}{\mu _{I_{s}}^{2}}+1 \right )\epsilon _{s}^{2}
\end{aligned}
\label{eq:proof25}
\end{equation}

\begin{equation}
\begin{aligned}
f(\overline{\omega}_{c}^{t+1}) &\leqslant f(\omega_{c}^{t})
-\left ( \frac{\eta _{c}}{2}+2\eta _{c}P_{c}^{2}B^{2}-\frac{LB^{2}(1+\gamma )^{2}}{\overline{\mu}^{2}P_{c}^{2}} \right )\left \| \bigtriangledown f(\omega _{c}^{t}) \right \|^{2}\\
&+\left (\frac{L(1+\gamma )^{2}}{\overline{\mu}^{2}P_{c}^{2}}\left ( \frac{4L^{2}P_{s}^{2}}{\mu _{I_{s}}^{2}}+1 \right )-\frac{8\eta _{c}L^{2}P_{c}^{2}P_{s}^{2}}{\mu _{I_{s}}^{2}}  \right )\epsilon _{s}^{2}-\eta _{c}\epsilon _{c}^{2}
\end{aligned}
\label{eq:proof26}
\end{equation}

At last, we get the convergence as:
\begin{equation}
\begin{aligned}
&\mathbb{E}_{s_{t}}\left [ f(\omega _{c}^{t+1}) \right ]\leqslant f(\omega_{c}^{t})- (\frac{\eta _{c}}{2}+2\eta _{c}P_{c}^{2}B^{2}-\frac{LB^{2}(1+\gamma )^{2}}{\overline{\mu}^{2}P_{c}^{2}}
-\frac{2(1+\gamma )B}{\overline{\mu}P_{c}\sqrt{K}}- \frac{2(2\sqrt{2K}+2)L(1+\gamma )^{2}B^{2}}{K\overline{\mu}^{2}P_{c}^{2}})\left \| \bigtriangledown f(\omega _{c}^{t}) \right \|^{2}\\
&+(\frac{\sqrt{K}\overline{\mu}P_{c}(1+\gamma )+BL(K+4\sqrt{2K}+4)(1+\gamma )^{2}}{KB\overline{\mu}^{2}P_{c}^{2}}
\left ( \frac{4L^{2}P_{s}^{2}}{\mu _{I_{s}}^{2}}+1 \right )-\frac{8\eta _{c}L^{2}P_{c}^{2}P_{s}^{2}}{\mu _{I_{s}}^{2}})\epsilon _{s}^{2}-\eta _{c}\epsilon _{c}^{2}
\end{aligned}
\label{eq:proof27}
\end{equation}

\end{proof}

\subsection{Cross-entropy Loss Functions}
With the optimization purpose changing, and the introduction of invariant aggregation and diversity transferring, the loss function of local clients also has huge changes. The optimization purpose of local client is to minimize the loss of each client $k\in \left | K \right |$.
\begin{equation}
\begin{aligned}
\underset{ \theta_{k}}{\mathit{\argmin}}\mathbb{E}_{\left ( x_{k},y_{k} \right )\in D_{k}^{*}}\left [ \mathfrak{h}_{k}\left ( x_{k},y_{k},\mathit{E}\right ) \right ]
\end{aligned}
\label{eq:loss}
\end{equation} 

where $D_{k}^{*}$ is the data distribution of client $k$, $D_{i}^{*}\neq D_{j}^{*},1\leqslant  i\neq j\leqslant K$ shows the Non-i.i.d factors between domains. In reality, the empirical loss function is employed for instead in actual calculations as:
\begin{equation}
\begin{aligned}
\hat{h}_{k}=\frac{1}{n_{k}}\sum_{j=1}^{n_{k}} \mathfrak{h}_{k}\left ( x_{k},y_{k}, \omega_{k} \right  ) 
\end{aligned}
\label{eq:empirical_loss}
\end{equation}
The training samples of $k$ client are defined as $D_{k}=\left \{ \left ( x_{k}^{j},y_{k}^{j} \right ),x_{k}^{j}\in \mathbb{R}^{d},y_{k}^{j}\in \mathbb{R} \right \}_{j=1}^{n_{k}}$, $n_{k}$ is the training sample number of client $k$.
With the application of the two-branch local model replacing single-branch, the local empirical loss function tends to:
\begin{equation}
\begin{aligned}
\widehat{h}_{k}=\frac{1}{n_{k}}\sum_{j=1}^{n_{k}}\mathfrak{h}_{k}\left ( P^{k}\left ( E_{c}^{k}(x_{k}^{j})\bigoplus E_{s}^{k}(x_{k}^{j}) \right ),y_{k}^{j} \right )
\end{aligned}
\label{eq:detail empirical loss}
\end{equation}



\end{document}